\documentclass[conference]{IEEEtran}

\usepackage{cite}
\usepackage{graphicx}
\usepackage{amsmath}
\usepackage{amssymb}
\usepackage{algorithm}
\usepackage{algpseudocode}
\usepackage{float}
\usepackage{booktabs}
\usepackage[table]{xcolor}
\usepackage{array}
\usepackage{multirow}
\ifCLASSOPTIONcompsoc
 \usepackage[caption=false,font=normalsize,labelfont=sf,textfont=sf]{subfig}
\else
 \usepackage[caption=false,font=footnotesize]{subfig}
\fi

\begin{document}

\title{Reinforcement Learning for Adaptive Traffic Signal Control: Turn-Based and Time-Based Approaches to Reduce Congestion}

\author{    
    \IEEEauthorblockN{Muhammad Tahir Rafique}
    \IEEEauthorblockA{\begin{minipage}{2in}\centering
    \textit{Department of Robotics \& Intelligent Machines Engineering}\\
    \textit{National University of\\Sciences \& Technology}\\
    Islamabad, Pakistan\\
    mragique.rime17smme@\\student.nust.edu.pk
    \end{minipage}}

    \and
    \IEEEauthorblockN{Ahmed Mustafa}
    \IEEEauthorblockA{\begin{minipage}{2in}\centering
    \textit{Department of Robotics \& Intelligent Machines Engineering}\\
    \textit{National University of\\Sciences \& Technology}\\
    Islamabad, Pakistan\\
    amustafa.rime20smme@\\student.nust.edu.pk
    \end{minipage}}

    \and
    \IEEEauthorblockN{Hasan Sajid}
    \IEEEauthorblockA{\begin{minipage}{2in}\centering
    \textit{Department of Robotics \& Intelligent Machines Engineering}\\
    \textit{National University of\\Sciences \& Technology}\\
    Islamabad, Pakistan\\
    hasan.sajid@smme.nust.edu.pk
    \end{minipage}}
}

\maketitle

\begin{abstract}
The growing demand for road use in urban areas has led to significant traffic congestion, posing challenges that are costly to mitigate through infrastructure expansion alone. As an alternative, optimizing existing traffic management systems, particularly through adaptive traffic signal control, offers a promising solution. This paper explores the use of Reinforcement Learning (RL) to enhance traffic signal operations at intersections, aiming to reduce congestion without extensive sensor networks. We introduce two RL-based algorithms: a turn-based agent, which dynamically prioritizes traffic signals based on real-time queue lengths, and a time-based agent, which adjusts signal phase durations according to traffic conditions while following a fixed phase cycle. By representing the state as a scalar queue length, our approach simplifies the learning process and lowers deployment costs. The algorithms were tested in four distinct traffic scenarios using seven evaluation metrics to comprehensively assess performance. Simulation results demonstrate that both RL algorithms significantly outperform conventional traffic signal control systems, highlighting their potential to improve urban traffic flow efficiently.
\end{abstract}

\hfill

\section{Introduction}
The rapid urbanization and population growth in cities worldwide have significantly increased the number of vehicles on roads, resulting in severe traffic congestion. This congestion not only aggravates fuel consumption and CO\textsubscript{2} emissions but also leads to substantial time loss for commuters, thereby impacting economic productivity and quality of life. Expanding road infrastructure to accommodate the growing traffic demand is often not a feasible solution due to the considerable financial burden it imposes on national economies. Consequently, there is a pressing need for more cost-effective strategies to manage traffic flow, particularly at intersections where congestion is most acute.

Traditional traffic signal control systems typically operate on fixed-time rotations for each direction at an intersection, without accounting for real-time traffic conditions. Such fixed-timing approaches can lead to significant inefficiencies. For example, signals may remain green for directions with little to no traffic, while vehicles on busier routes experience unnecessary delays. During peak hours or special events, like sporting events or school dismissals, these systems often become overwhelmed, resulting in severe congestion. In some cases, traffic police officers manually adjust signal timings based on real-time observations; however, these adjustments are inherently limited by their localized perspective and inability to optimize traffic flow on a broader scale.

In developing countries, where road infrastructure is rapidly evolving and traffic patterns are constantly changing, there is an urgent need for adaptive traffic management systems that can dynamically adjust to fluctuating traffic conditions. Recent advancements in machine learning, particularly in Reinforcement Learning (RL), offer promising solutions for developing such adaptive control systems. RL has been successfully applied in various domains, including games \cite{playingAtari}, robotics \cite{learningHandEye}, healthcare \cite{medicalEg}, finance \cite{financeEg}, and traffic signal control \cite{rlBook}. Unlike traditional traffic signal systems, RL-based approaches do not require a perfect model of the environment. Instead, they learn optimal strategies through interaction with the environment, leveraging a trial-and-error approach to maximize long-term rewards.

Despite the potential of RL for traffic signal control, its real-world application has been limited, primarily due to the high costs associated with deploying state-detection sensors \cite{trafficSignalTimingVia} \cite{intellilight}. In many developing countries, these costs are prohibitive, and alternative methods are needed. One viable solution is to utilize existing camera-based surveillance systems, which are commonly installed at intersections for security purposes. These systems can be integrated with object detection algorithms to count stationary vehicles, providing the necessary state information for RL agents at a fraction of the cost of traditional sensor networks.

This paper proposes an RL-based framework for adaptive traffic signal control aimed at improving traffic flow at intersections. Our approach introduces a novel state representation using a scalar measure of queue length, which simplifies the learning process and reduces the reliance on extensive sensor networks. This research differs from previous studies \cite{trafficSignalTimingVia} \cite{usingADeepLearnig} \cite{trafficLightControl} by employing a comprehensive set of seven evaluation metrics to thoroughly assess the performance of two RL algorithms—a turn-based agent and a time-based agent—across four distinct traffic scenarios.

The remainder of this paper is organized as follows: Section II reviews related work in the field of RL-based traffic signal control. Section III formulates the traffic signal control problem within an RL framework. Section IV details the evaluation metrics and presents simulation results, and Section V concludes the study with insights and future research directions.
\section{Related Work}
Traffic flow optimization is a critical research area in transportation systems, with ongoing efforts spanning several decades. Early work in this field, such as that by Webster \cite{roadResearch}, utilized mathematical approaches to determine optimal traffic signal phase times. Webster's initial model was later refined by Robertson \cite{optimizingNetwok} to accommodate increasing traffic volumes at intersections, highlighting the evolving nature of traffic signal control methodologies.

With advancements in sensor and computer technology, more sophisticated techniques such as fuzzy control \cite{fuzzyControl, fuzzyModel} and genetic algorithms \cite{geneticAlgo, hybridGenetic} have been introduced to traffic control systems. These methods typically rely on historical data to predict traffic patterns, but their performance diminishes as traffic behaviors change over time. In recent years, reinforcement learning (RL) has gained significant attention due to its adaptive learning capabilities \cite{adaptiveTraffic, rlApplication}. A comprehensive review by El-Tantawy \cite{designRL} summarized RL approaches from 1997 to 2010, focusing on methods that utilized Q-tables \cite{deepRlBook} and linear functions for calculating Q-values.

Li \textit{et al.} \cite{trafficSignalTimingVia} proposed using a deep stacked autoencoder (SAE) network to determine actions based on the current state of traffic. Their algorithm takes queue length as input and uses the difference between north-south and east-west traffic as a reward signal. Although their approach demonstrated a 14\% reduction in average traffic delay compared to conventional traffic light control, it lacked a detailed explanation of network parameter updates, which are crucial for algorithm stability.

Genders \textit{et al.} \cite{usingADeepLearnig} introduced a deep reinforcement learning algorithm employing a convolutional neural network to optimize Q-values. Their state representation combined a vehicle position matrix, vehicle velocity matrix, and the most recent state of the intersection. The vehicle position matrix was created by segmenting the road into small sections, indicating vehicle presence, while the vehicle velocity matrix normalized speeds against speed limits. Although this comprehensive state representation proved effective in simulations, it presents practical challenges for real-world implementation due to its complexity. The authors utilized four evaluation metrics and reported superior performance on three of them. However, the scalar representation of queue length, averaged across all intersection sides, can obscure performance issues in scenarios where traffic distribution is uneven.

Gao \textit{et al.} \cite{adaptiveTrafficSignal} proposed an RL algorithm that automatically extracts features from raw traffic data to learn optimal policies. The algorithm defined actions to control signals for either east-west or north-south traffic flows, but did not account for left or right turns, limiting its applicability to specific traffic conditions. In simulations with varied traffic distributions, the algorithm showed improved performance over conventional traffic light systems by minimizing vehicle delay times.

Recent research has primarily focused on optimizing signal phase selection, with less attention given to phase duration. Liang \textit{et al.} \cite{Liang_2019} addressed this gap by adjusting signal durations dynamically based on data from vehicular networks. Their approach utilized a dueling network architecture, double Q-learning, and prioritized experience replay to adapt signal timings. State information was captured using square-shaped grids at intersections, where each vehicle's presence was registered within these grids. However, this method could lead to inaccuracies in real-world scenarios where vehicles vary in size, potentially spanning multiple grid cells. The authors defined actions as incremental adjustments to phase durations, which remained constant for a full cycle, suggesting potential improvements by tailoring durations based on real-time traffic conditions for each side of the intersection.

Some researchers \cite{intellilight, trafficLightControl} have explored using raw features rather than human-crafted features to define state representations. For instance, states have been represented by stacking images of intersection snapshots, capturing the motion through a sequence of frames. While this approach could theoretically be implemented using overhead cameras or drones, practical deployment poses significant logistical challenges.

In contrast to the existing literature, our study introduces an algorithm designed for real-world applicability by employing a state representation based on scalar queue length. We evaluate the performance of our algorithm using seven distinct metrics, offering a more comprehensive assessment than previous studies that focused either on phase selection or phase duration alone. By integrating both aspects, our approach provides a balanced and practical solution for adaptive traffic signal control.
\section{System Design}
\subsection{State Representation}
The agent's state represents the environmental conditions at a specific sampling interval, $ts$ , and is denoted by $s_{ts}$. The design of the state is crucial in reinforcement learning, as it significantly impacts the agent's performance. In this study, the state is characterized by the queue lengths at an intersection. Queue length, a scalar quantity, is determined by counting the total number of stationary vehicles at the intersection. For a turn-based agent, the state is defined by obtaining queue lengths from all approaches of the intersection. The number of queue lengths required depends on the configuration of the intersection; for instance, a four-way intersection necessitates four queue lengths to define the state. In contrast, for a time-based agent, the state is represented by the queue length of a single approach to the intersection.

\begin{figure}
    \centering
    \includegraphics[scale=1.8]{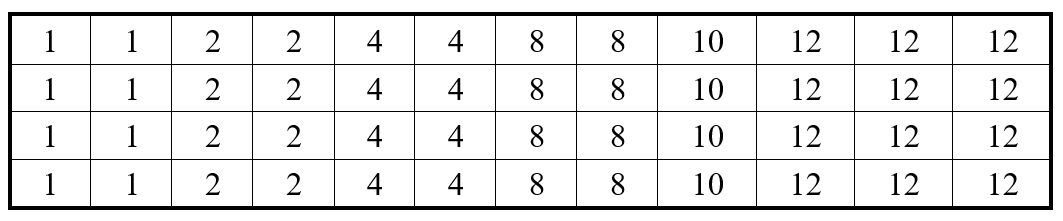}
    \vspace{-3mm}
    \caption{48-bit encoding matrix used to represent queue lengths, capable of encoding up to 304 stationary vehicles at an intersection}
    \label{fig:state}
    \vspace{-2mm}
\end{figure}

In the simulation environment, the state is determined by counting all stationary vehicles at the intersection, where vehicles moving at speeds below $1 m/s$ are considered stationary. The resulting state is a scalar number, which is not directly suitable for training a reinforcement learning agent due to its low dimensionality. To effectively model the relationship between state and action, a deep neural network is utilized. However, training such a network using scalar inputs is impractical, as the low input dimensionality prevents effective learning. To address this issue, we employ human-crafted features to enhance the input dimensionality and facilitate training. In this research, binary encoding is utilized to increase the state’s dimensionality. The encoding process involves two key parameters: encoding size and encoding weights. Encoding size specifies the number of cells in the encoding matrix, while encoding weights determine the minimum number of vehicles required to populate a cell with a $1$ in the encoding matrix. The procedure for filling the encoding matrix is outlined in Algorithm \ref{alg:state}, which provides the pseudocode detailing the steps involved. Fig. \ref{fig:state} illustrates the encoding matrix, along with the encoding weights assigned to each cell.

The encoding matrix is designed to accommodate up to $304$ vehicles per approach at an intersection. When the queue length for an approach reaches $304$ vehicles, every cell in the corresponding encoding matrix will be filled with $1$s. Conversely, if no vehicles are present, all cells in the encoding matrix will be set to $0$s. 
For turn-based agents, encoding matrices are generated for all approaches of the intersection and then concatenated to form a single vector representing the state of the intersection. This vector can be used to train the reinforcement learning agent effectively.

 \begin{algorithm}[b]
    \caption{Get State Encoding.}
    \label{alg:state}
    \begin{algorithmic}[1]
      \Function{GetStateEncoding}{}
        \State Initialize the state matrix with zeros.
        \State ${M} \xleftarrow[] {} zeros(4, 12)$
        \State Initialize the encoding weights matrix.
        \State ${E} \xleftarrow[] {} EncodingWeights(4, 12)$
        \State Find the queue length
        \State ${ql} \xleftarrow[] {} Queue $
        \For{$j = 1$ to ${M.columns}$}
            \For{$i = 1$ to ${M.rows}$}
                \State $C \xleftarrow[] {} E[i, j]$
                \If{$ql \geq C$}
                    \State $ql = ql - C$
                    \State $M[i, j] = 1$
                \EndIf
            \EndFor
        \EndFor
        \State return $M$
       \EndFunction
\end{algorithmic}
\end{algorithm}

\subsection{Action Space}
To effectively navigate vehicles through an intersection, the reinforcement learning agent must select an appropriate action based on the current state. For a turn-based agent, an action corresponds to the possible traffic phase, while for a time-based agent, the action represents the duration of the green light phase.

The potential actions for a turn-based agent are defined in \eqref{eq:turn_action}. The action \textit{NG} (North-Green) allows vehicles to proceed from the north to other directions (East, West, and South), signaling vehicles on the East, West, and South routes to stop and not enter the intersection. Similarly, \textit{WG} (West-Green), \textit{EG} (East-Green), and \textit{SG} (South-Green) permit vehicles from the West, East, and South, respectively, to proceed to other directions while preventing the movement of vehicles from the remaining directions through the intersection. Specifically, for \textit{WG} vehicles on the North, East, and South routes must stop; for \textit{EG} those on the North, West, and South routes must stop; and for \textit{SG} vehicles on the East, West, and North routes must stop.
\begin{equation}
    A_{turn}=\{NG,\ WG,\ EG,\ SG\}
    \label{eq:turn_action}
\end{equation}

In contrast, a time-based agent does not select a phase but rather the duration for which a phase remains active. The phase cycle remains fixed, but the duration of each green light phase is adjusted based on the environmental state. The possible actions for a time-based agent are given in \eqref{eq:time_action}. Here, a value of $0$ indicates no change to the base duration of the green light, while a value of $10$ signifies an addition of ten seconds to the base green light duration. Given a base green light time of $15$ seconds, the minimum duration for any phase is $15$ seconds, and the maximum is $34$ seconds.
\begin{equation}
    \begin{split}
        A_{time}=\{&0,\ 1,\ 2,\ 3,\ 4,\ 5,\ 6,\ 7,\ 8,\ 9,\ 10\\ &11, \ 12, \ 13, \ 14, \ 15, \ 16, \ 17, \ 18, \ 19 \}
    \end{split}
    \label{eq:time_action}
\end{equation}

\begin{figure}[t]
    \vspace{-2mm}
    \centering
    \includegraphics[scale=1.5]{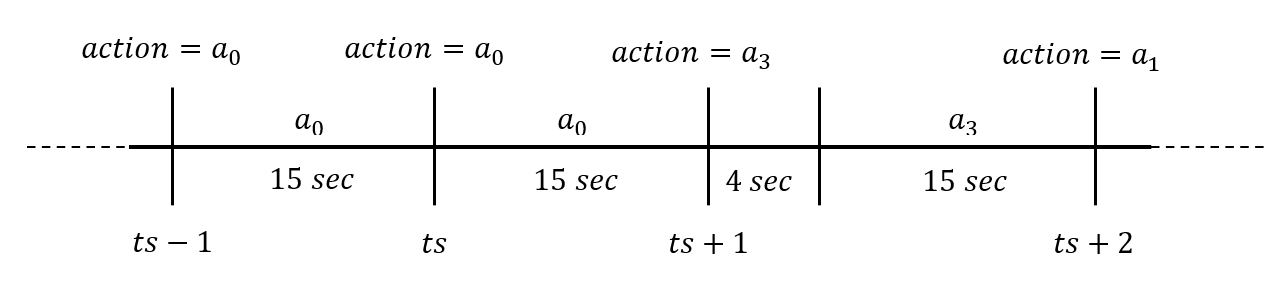}
    \vspace{-4mm}
    \caption{Action timeline for turn-based agent}
    \vspace{-3mm}
    \label{fig:turn_action}
\end{figure}

The reinforcement learning agent selects an action at each sampling time step $ts$. Once an action is chosen, the agent will select the next action at the subsequent time step $ts+1$ In a conventional traffic light control system, a transition phase occurs after each green phase, during which the yellow light is activated for four seconds. For a turn-based agent, this transition phase only occurs if the selected phase differs from the previous phase. For example, if the action selected by the agent at time step $ts$ is $a_0$ and is the same as the action taken at the previous time step $ts-1$, no transition phase will occur. After the completion of action $a_0$, the agent selects a new action at the next time step $ts+1$. If, at this point, the agent selects action $a_3$, which is different from the previous action, a transition phase of four seconds will occur. The timeline of the turn-based agent's actions is illustrated in Fig. \ref{fig:turn_action}.

\begin{figure}[t]
    \centering
    \includegraphics[scale=1.5]{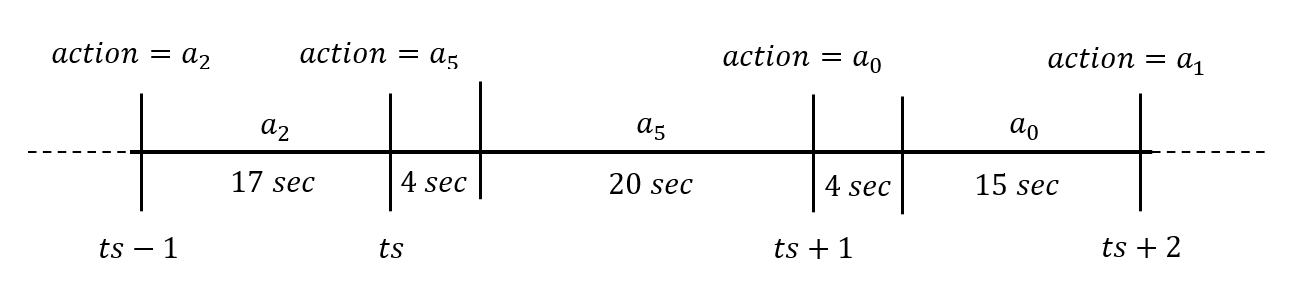}
    \vspace{-4mm}
    \caption{Action timeline for time-based agent}
    \vspace{-1mm}
    \label{fig:time_action}
\end{figure}

For a time-based agent, the transition phase occurs consistently after every green phase, mirroring the behavior of a conventional traffic light control system. The timeline for the actions of a time-based agent is depicted in Fig. \ref{fig:time_action}.

 

\subsection{Reward}
The reward serves as a quantitative measure of feedback that the reinforcement learning agent receives from the environment following the execution of a particular action. This feedback is crucial for the agent to evaluate the efficacy of its action, determining whether it had a favorable or adverse impact on the traffic flow. The agent utilizes this information to refine its policy, thereby optimizing future action selections. The reward can assume either a positive or negative value, where a positive reward signifies that the selected action has effectively alleviated traffic congestion, whereas a negative reward indicates a deterioration in traffic conditions.

In this research, our central objective is to optimize intersection efficiency by reducing vehicle wait times. To achieve this, the reward is defined as the difference in accumulative wait time ($awt$) of all vehicles across two consecutive sampling intervals
\begin{equation}
    r_{(ts)}=\ {awt}_{(ts-1)}-\ {awt}_{(ts)}
    \label{eq:reward}
\end{equation}

where $awt$ represents the accumulative wait time, calculated by summing the wait times of all stationary vehicles at all sides of the intersection. The reward $r_{ts}$ is computed as the difference between the accumulative wait times at the previous timestep $awt_{ts-1}$ and the current timestep ${awt}_{(ts)}$. Equation \eqref{eq:reward} is designed such that a positive reward is given when an agent takes a beneficial action, and a negative reward is given when an action is detrimental.

\subsection{Policy}
The policy is a function that maps the current state of the environment to a corresponding action. To achieve optimal wait time, the agent observes the current state, selects an action according to policy $\pi$, and receives a reward. The agent aims to maximize the immediate reward $r_{(ts)}$ by choosing actions that yield the highest immediate benefit. The optimal policy ${\pi }^*$ is defined as the policy that maximizes the cumulative reward, given by:
\begin{equation} \label{eq:acc_reward}
    \begin{split}
    Q_{\pi }\left(s,a\right) & = \mathbb{E}[r_{ts}+\ \gamma r_{ts+1}+{\gamma }^2r_{ts+2}+\dots ] \\
     & = \mathbb{E}\left[\sum^{\infty }_{k=0}{{\gamma }^kr_{ts+k}}\right]
    \end{split}
\end{equation}

where $\gamma$ is a discount factor in the range $0 \leq \gamma \leq 1$, reflecting the importance the agent places on future rewards.

To determine the policy, we employ a deep neural network as a function approximator, parameterized by $\theta$. The policy distribution, $\pi (a_{ts}|s_{ts};\theta )$, is optimized via experience replay. At each sampling time step, the experience tuple $(s_{ts},a_{ts},r_{ts},s_{ts+1})$ is stored in a replay buffer. Following several episodes of simulation, a substantial dataset is accumulated to train the neural network effectively. A mini-batch of $64$ samples is periodically drawn from this buffer to update the network parameters. After sufficient training, the parameters converge, yielding the optimal policy $\pi \left(a\mathrel{\left|\vphantom{a s;{\theta }^*}\right.\kern-\nulldelimiterspace}s;{\theta }^*\right)$ as represented by:
\begin{equation} \label{eq:opt_policy}
    \pi \left(a\mathrel{\left|\vphantom{a s;{\theta }^*}\right.\kern-\nulldelimiterspace}s;{\theta }^*\right)={\mathrm{arg} {\mathop{max}_{\pi } Q\left(s,a\right)\ \ }\ }\ \ \ \forall \ s\in S,\ a\in A
\end{equation}
\section{Results}
\subsection{Experimental setup}
The Simulation of Urban MObility (SUMO) \cite{sumo} traffic simulator was employed in our experiments. SUMO, an open-source microscopic traffic simulator, provides comprehensive data on vehicles and network conditions. The simulation settings are detailed in the subsequent sections.

\subsubsection{Intersection} 
\begin{figure}
    \centering
    \includegraphics[scale=0.8]{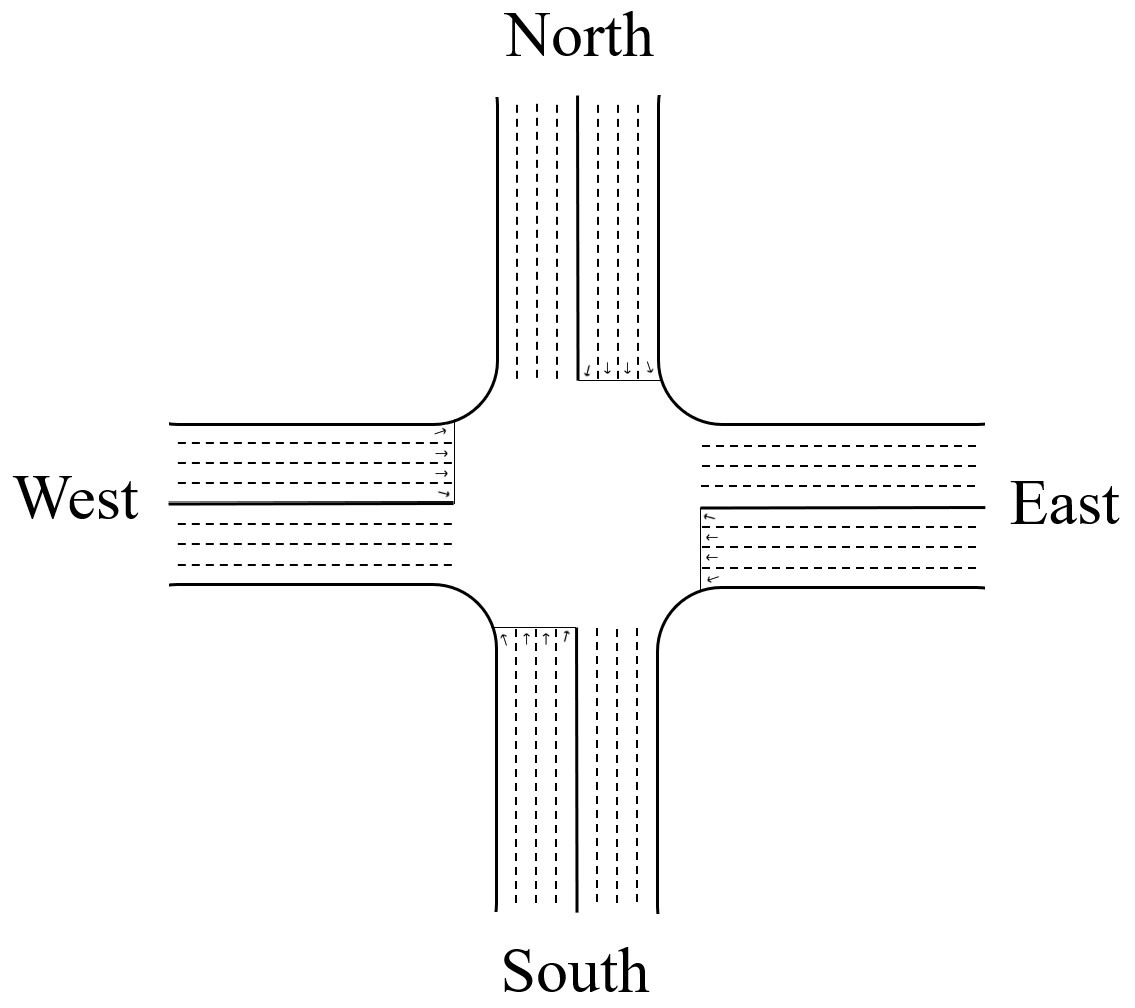}
    \vspace{-3mm}
    \caption{Four-way intersection}
    \label{fig:intersection}
\end{figure}

The intersection geometry utilized in our study is depicted in Fig. \ref{fig:intersection}. It features a four-way layout with four incoming and four outgoing roads, each having four lanes. The simulation assumes a left-hand driving rule: the left-most lane permits left turns, the middle two lanes are designated for through traffic, and the right-most lane is for right turns. The approach and departure roads are each set to 750 meters in length. Vehicle attributes used in the simulation are provided in Table \ref{tab:attribute}.

\subsubsection{Phase}
The traffic light system in this experiment incorporates eight distinct phases, as outlined in Table \ref{tab:phase}. Both the conventional traffic light control and the time-based agent cycle through these phases sequentially, completing a full cycle after the $8^{th}$ phase before returning to the $1^{st}$ phase. In contrast, the turn-based agent dynamically selects phases based on real-time traffic conditions, specifically controlling phases $1$, $3$, $5$, and $7$. The phase duration, defined as the active period of a phase at the intersection, remains constant for the conventional control and turn-based agent but is variable for the time-based agent, which can adjust the duration for phases $1$, $3$, $5$, and $7$.

\subsubsection{Traffic Generation}
Traffic is generated randomly using a Weibull distribution to reflect realistic traffic patterns, where flow rates typically increase rapidly and decrease slowly. Four distinct traffic scenarios are simulated: low, high, east-west (\textit{EW}), and north-south (\textit{NS}). In the low and high traffic scenarios, an equal number of vehicles are introduced from all directions. In the \textit{EW} scenario, a higher volume of traffic is generated on the East and West approaches, while the \textit{NS} scenario prioritizes the North and South approaches. The specific vehicle counts for each scenario are provided in Table \ref{tab:traffic_gen}. Of the generated vehicles, $60\%$ proceed straight, while the remaining $40\%$ make left or right turns. Once the origin and destination of each vehicle are determined, the $A^*$ path planning algorithm is employed to compute the optimal route.

\begin{table}[t]
    \vspace{-3mm}
    \centering
    \caption{Vehicle attributes}
    \vspace{-2mm}
    \begin{tabular}{>{\centering}p{3cm} >{\centering\arraybackslash}p{3cm}}
    \toprule
    \textbf{Attributes} & \textbf{Values} \\ 
    \midrule
    Length \cellcolor[HTML]{f2f2f2} &  $5\ m$ \\
    Width \cellcolor[HTML]{f2f2f2} &  $1.8\ m$ \\
    Minimum gap \cellcolor[HTML]{f2f2f2} & $2.5\ m$ \\
    Maximum speed \cellcolor[HTML]{f2f2f2} & $25{m}/{s}$   \\
    Maximum acceleration \cellcolor[HTML]{f2f2f2} & $1{m}/{s^2}$   \\ 
    Maximum deceleration \cellcolor[HTML]{f2f2f2} & $4.5{m}/{s^2}$   \\ 
    \bottomrule
    \end{tabular}
    \label{tab:attribute}
    \vspace{-1mm}
\end{table}
\begin{table}[b]
    \vspace{-2mm}
    \centering
    \caption{Four-way intersection phases}
    \vspace{-2mm}
    \begin{tabular}{>{\centering}p{4em} c c c c >{\centering\arraybackslash}p{7em}}
    \toprule
    \textbf{Phase} & \multirow{2}{*}{\textbf{North}} & \multirow{2}{*}{\textbf{West}} & \multirow{2}{*}{\textbf{East}} & \multirow{2}{*}{\textbf{South}} & \textbf{Phase Duration} \\
    \textbf{Number} & & & & & \textbf{(Seconds)}\\
    \midrule
    1\cellcolor[HTML]{F2F2F2} & g\cellcolor[HTML]{EAF1DD} & r\cellcolor[HTML]{F2DBDB} & r\cellcolor[HTML]{F2DBDB} & r\cellcolor[HTML]{F2DBDB} & 15\cellcolor[HTML]{F2F2F2} \\
    2\cellcolor[HTML]{F2F2F2} & y\cellcolor[HTML]{FFFFCC} & r\cellcolor[HTML]{F2DBDB} & r\cellcolor[HTML]{F2DBDB} & r\cellcolor[HTML]{F2DBDB} & 4\cellcolor[HTML]{F2F2F2} \\
    3\cellcolor[HTML]{F2F2F2} & r\cellcolor[HTML]{F2DBDB} & g\cellcolor[HTML]{EAF1DD} & r\cellcolor[HTML]{F2DBDB} & r\cellcolor[HTML]{F2DBDB} & 15\cellcolor[HTML]{F2F2F2} \\ 
    4\cellcolor[HTML]{F2F2F2} & r\cellcolor[HTML]{F2DBDB} & y\cellcolor[HTML]{FFFFCC} & r\cellcolor[HTML]{F2DBDB} & r\cellcolor[HTML]{F2DBDB} & 4\cellcolor[HTML]{F2F2F2} \\
    5\cellcolor[HTML]{F2F2F2} & r\cellcolor[HTML]{F2DBDB} & r\cellcolor[HTML]{F2DBDB} & g\cellcolor[HTML]{EAF1DD} & r\cellcolor[HTML]{F2DBDB} & 15\cellcolor[HTML]{F2F2F2} \\
    6\cellcolor[HTML]{F2F2F2} & r\cellcolor[HTML]{F2DBDB} & r\cellcolor[HTML]{F2DBDB} & y\cellcolor[HTML]{FFFFCC} & r\cellcolor[HTML]{F2DBDB} & 4\cellcolor[HTML]{F2F2F2} \\
    7\cellcolor[HTML]{F2F2F2} & r\cellcolor[HTML]{F2DBDB} & r\cellcolor[HTML]{F2DBDB} & r\cellcolor[HTML]{F2DBDB} & g\cellcolor[HTML]{EAF1DD} & 15\cellcolor[HTML]{F2F2F2} \\
    8\cellcolor[HTML]{F2F2F2} & r\cellcolor[HTML]{F2DBDB} & r\cellcolor[HTML]{F2DBDB} & r\cellcolor[HTML]{F2DBDB} & y\cellcolor[HTML]{FFFFCC} & 4\cellcolor[HTML]{F2F2F2} \\
    
    \bottomrule
    \end{tabular}
    \label{tab:phase}
    \vspace{-5mm}
\end{table}
\begin{table}[t]
    \vspace{-3mm}
    \centering
    \caption{Number of Generated vehicles}
    \vspace{-2mm}
    \begin{tabular}{>{\centering}p{3cm} >{\centering\arraybackslash}p{3cm}}
    \toprule
    \textbf{Traffic Scenario} & \textbf{Generated Vehicles}\\
    \midrule
    Low\cellcolor[HTML]{f2f2f2} & 600 \\
    High\cellcolor[HTML]{f2f2f2} & 3000 \\
    East-West (EW)\cellcolor[HTML]{f2f2f2} & 1500 \\
    North-South (NS)\cellcolor[HTML]{f2f2f2} & 1500 \\
    \bottomrule
    \end{tabular}
    \label{tab:traffic_gen}
    \vspace{-1mm}
\end{table}
\subsection{Neural Network architecture}
\begin{figure}[b]
    \vspace{-2mm}
    \centering
    \includegraphics[scale=1.3]{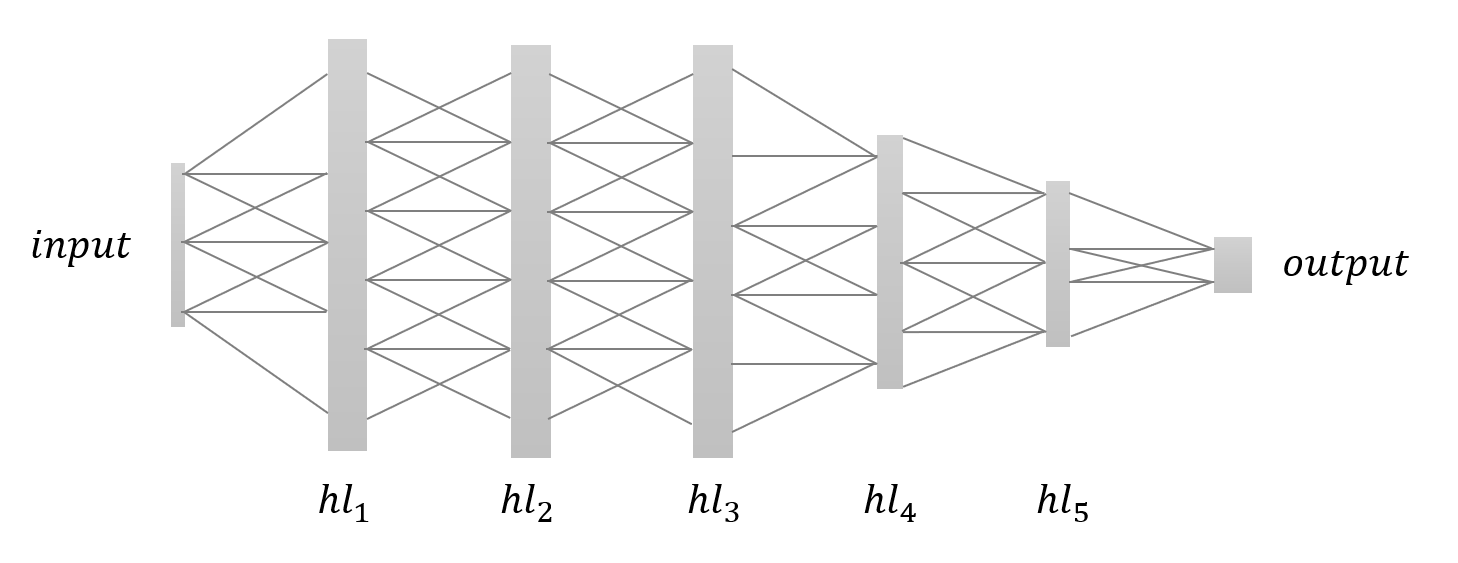}
    \vspace{-3mm}
    \caption{DNN Architecture}
    \label{fig:deep_nn}
    \vspace{-5mm}
\end{figure}

The RL agents utilize a deep neural network as a function approximator to model the state-action value function. The neural network architecture comprises $5$ fully connected hidden layers, denoted by $hl$, each employing rectified linear units (ReLU) as activation functions. The architecture details are depicted in Fig. \ref{fig:deep_nn}. Specifically, hidden layers ${hl}_1$, ${hl}_2$, ${hl}_3$, ${hl}_4$ and ${hl}_5$ contain $512$, $512$, $512$, $256$, and $128$ nodes, respectively. 

The network includes an input layer and an output layer, where the input layer receives the state representation of the environment, and the output layer computes the Q-values for the available actions. The dimensionality of both input and output layers is contingent on the agent type. For a turn-based agent, the number of input nodes corresponds to the product of the encoding matrix size and the number of intersection approaches, while the output layer nodes align with the size of the action space $A_{turn}$. Conversely, for a time-based agent, the input layer size matches the encoding matrix, and the output layer size corresponds to the action space $A_{turn}$ The output layer is fully connected with a linear activation function.
\subsection{Performance Metrics}
To rigorously evaluate the performance of the reinforcement learning agent, multiple metrics are employed rather than relying on a single measure. This multi-faceted approach provides a comprehensive assessment of the agent's ability to manage traffic efficiently under various conditions. The following sections outline the specific metrics used in this study.

\subsubsection{Total Negative Reward}
The Total Negative Reward (\(Tnr\)) is a cumulative measure that quantifies the penalties incurred by the agent due to actions that exacerbate traffic congestion. Unlike positive rewards that signal desirable actions, negative rewards penalize the agent for decisions that lead to increased wait times or longer queues. \(Tnr\) is calculated by summing all negative rewards received during a single episode of simulation, emphasizing the importance of minimizing detrimental actions:
\begin{equation}
    \label{tnr}
    {Tnr}_{(e)}=~\sum^m_{ts=1}{{\mathrm{min}(0,\ r}_{(ts)})}
\end{equation}

where \(ts\) denotes the sampling time step, \(m\) is the total number of sampling time steps within an episode, \(e\) represents the episode index, and \(r_{(ts)}\) is the reward at time step \(ts\). By focusing on the accumulation of negative rewards, the agent is incentivized to adopt strategies that reduce traffic congestion.

\subsubsection{Total Accumulative Wait Time}
Total Accumulative Wait Time (\(Tawt\)) is another critical metric that captures the aggregate waiting time of all vehicles across an episode. It is defined as the sum of individual vehicle wait times (\(wt\)) on the incoming road segments at each sampling time step. The accumulative wait time at a specific time step \(ts\) is given by:
\begin{equation}
    \label{awt_ts}
    {awt}_{(ts)}=~\sum^n_{c=1}{\left[{rd}^{(c)}~{wt}^{(c)}_{(ts)}\right]}
\end{equation}

where \(n\) is the total number of vehicles in the network, and \({rd}^{(c)}\) indicates whether a vehicle \(c\) is on an incoming road:

\[
{rd}^{(c)}= 
\begin{cases} 
1 & \text{if the vehicle is on an incoming road} \\
0 & \text{otherwise} 
\end{cases}
\]

The total accumulative wait time for an episode, \(Tawt_{(e)}\), is then the sum of all \({awt}_{(ts)}\) over the course of the episode:
\begin{equation}
    \label{eq:tawt}
    {Tawt}_{(e)}=~\sum^m_{ts=1}{{awt}_{(ts)}}
\end{equation}

A lower \(Tawt_{(e)}\) indicates improved traffic flow and reduced congestion.

\subsubsection{Expected Wait Time per Vehicle}
The Expected Wait Time per Vehicle (\(ewpv\)) provides an average measure of the waiting time that a vehicle experiences when passing through the intersection. It is calculated at the end of each episode by averaging the wait times of all vehicles:
\begin{equation}
    \label{eq:ewpv}
    {ewpv}_{(e)}=~\frac{\sum^n_{c=1}{{rd}^{(c)}~{wt}^{(c)}_{(ts=m)}}}{n}
\end{equation}

This metric is crucial for evaluating the efficiency of the agent; a lower \(ewpv_{(e)}\) reflects better performance in minimizing vehicle delays.

\subsubsection{Average Queue Length}
Average Queue Length (\(aql\)) is used to assess the typical number of stationary vehicles at the intersection, a direct indicator of congestion. Vehicles are considered stationary if their speed is less than 1 m/s. The queue length at each sampling time step \(ts\) is calculated as:

\[
{ql}_{(ts)}=\sum^n_{c=1}{{rd}^{(c)}~\left[1-\text{floor}\left(\frac{\text{sgn}(v)}{2}+1\right)\right]}
\]

where \(v = p^{(c)}_{(ts)} - 1\) represents the adjusted speed of vehicle \(c\) at time step \(ts\), and \(\text{sgn}\) is the signum function defined as:

\[
\text{sgn}(v) = 
\begin{cases} 
1 & \text{if } v > 0 \\
0 & \text{if } v = 0 \\
-1 & \text{if } v < 0 
\end{cases}
\]

The floor function rounds down its argument, ensuring that only vehicles with negligible speeds are counted as part of the queue. The average queue length over an episode \(e\), \({aql}_{(e)}\), is the mean of \(ql\) across all time steps:
\begin{equation}
    \label{eq:aql}
    {aql}_{(e)}=~\frac{\sum^m_{ts=1}{{ql}_{(ts)}}}{m}
\end{equation}

A lower \({aql}_{(e)}\) indicates effective traffic management and minimal congestion.
\subsection{Training}
Both turn-based and time-based agents were trained for $300$ episodes, each corresponding to $5400$ seconds of simulated traffic. Extending training beyond 300 episodes did not yield further performance improvements, indicating convergence. To avoid overfitting, traffic scenarios were varied after each episode.

During training, the agent seeks to optimize its policy by selecting actions that maximize cumulative rewards. Initially, agents explore the state-action space by choosing actions randomly. As training progresses, the agent transitions from exploration to exploitation, guided by the exploration rate, \(\varepsilon\). This rate, which controls the probability of taking exploratory actions, decreases over time following a non-linear decay function as defined in Equation \eqref{eq:exploration}.

\begin{equation}
    \varepsilon= \begin{cases} 
      1 & \text{~~~~~~~$e\leq90$} \\
      1-\frac{0.8}{120}(e-90) & \text{$90<e\leq210$} \\
      0.2-\frac{0.2}{90}(e-210) & \text{$210<e\leq300$}
   \end{cases}
    \label{eq:exploration}
\end{equation}

Equation \eqref{eq:exploration} maintains a high exploration rate for the first $90$ episodes to ensure comprehensive exploration. From episodes $90$ to $210$, \(\varepsilon\) decreases rapidly, shifting focus toward exploiting learned strategies. In the final phase, between episodes $210$ and $300$, \(\varepsilon\) decreases more gradually, allowing for occasional exploration to fine-tune the policy.

\begin{algorithm}
   \caption{Reinforcement learning algorithm with experience replay.}
    \begin{algorithmic}[1]
    \State Initialize the policy network parameters $\theta$ with random values.
    \State Initialize the hyper-parameters ($ \alpha, \gamma, \varepsilon, B, N$).
    \State Initialize traffic scenario $TS$ with (low, high, EW, NS) traffic.
    \State Initialize replay memory $M$ with size $L$.
    \For{$episode = 1$ to ${N}$}
        \State Generate Traffic according to traffic scenario $TS$
        \For{$t = 1$ to $steps$}
            \State ${s} \xleftarrow[] {} GetStateEncoding()$
            \State ${rand} \xleftarrow[] {} RandomNumberGenerator()$
            \If{$rand \leq \varepsilon$}
                \State ${a} \xleftarrow[] {} RandomAction()$
            \Else
                \State ${a} \xleftarrow[] {} arg\max\limits_{\pi} Q(s, a)$
            \EndIf
            \State ${r, s_{t+1}} \xleftarrow[] {} ApplyAction(a)$
            \State ${M} \xleftarrow[] {} Add(s, a, r, s_{t+1})$
            \If{$size(M) \geq B$}
                \State ${b} \xleftarrow[] {} RandomSample(M, B)$
                \For{$i = 1$ to $B$}
                    \If{$s^{(i)}_{t+1}$ is terminal}
                        \State $y^{(i)} \xleftarrow[] {} r^{(i)}$
                    \Else
                        \State $y^{(i)} \xleftarrow[] {} r^{(i)} + \gamma \max\limits_{a} Q(s^{(i)}_{t+1}, a)$
                    \EndIf
                \EndFor
                \State $J = \frac{1}{B}\sum_{j=1}^{B}(y^{(j)} - Q(s^{(j)}, a^{(j)}))^2$
                \State Update $\theta$ with $\Delta J$ using Adam Optimizer.
            \EndIf
            \State $s \xleftarrow[] {} s_{t+1}$
        \EndFor
        \State Update $\varepsilon$
        \State Update $TS$
    \EndFor
\end{algorithmic}
\label{alg:training}
\end{algorithm}

Training data, consisting of state, action, reward, and next state tuples, are stored in an experience replay buffer with a capacity of $50,000$ examples. The buffer is updated continuously, with older samples being replaced as new data is collected. At each sampling timestep, a batch of experiences is randomly sampled from the buffer to update the neural network parameters. 

\begin{figure*}
     \centering
     \subfloat[Turn-based agent negative reward\label{fig:turn_tnr}]{%
       \includegraphics[width=0.24\linewidth]{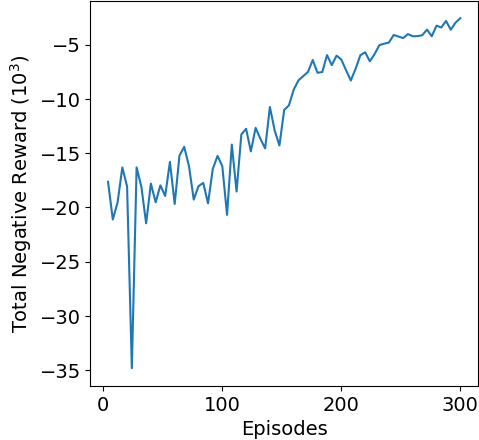}}
    \hfill
    \subfloat[Turn-based agent average queue length\label{fig:turn_aql}]{%
        \includegraphics[width=0.24\linewidth]{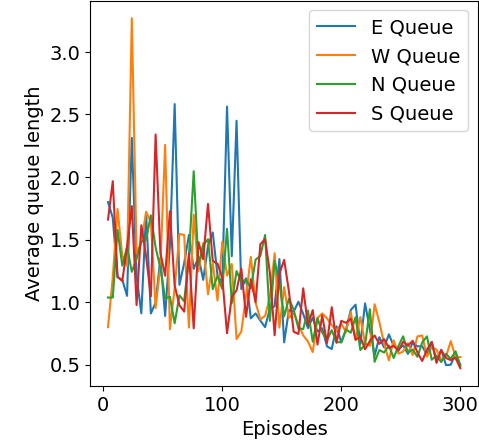}}
    \hfill
    \subfloat[Time-based agent negative reward\label{fig:time_tnr}]{%
       \includegraphics[width=0.24\linewidth]{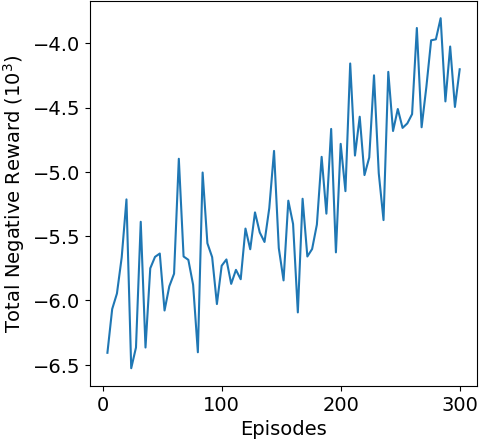}}
    \hfill
    \subfloat[Time-based agent average queue length\label{fig:time_aql}]{%
        \includegraphics[width=0.24\linewidth]{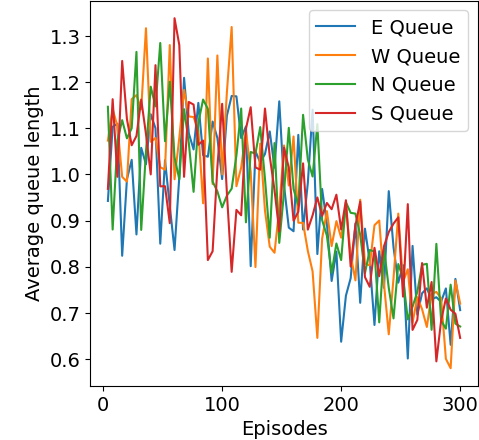}}
    \caption{Negative reward and averAage queue length while training.}
    \label{fig:trining}
\end{figure*}

The neural network inputs the current state and outputs Q-values for all possible actions. Target Q-values are computed using the Bellman equation and compared against the network's predictions to calculate the loss, which is minimized using the Adam optimizer. Hyperparameters such as the learning rate ($0.001$) and batch size ($64$) were chosen based on preliminary experiments. The full training process is outlined in Algorithm \ref{alg:training}.

figures \ref{fig:turn_tnr} and \ref{fig:turn_aql} show the training results for the turn-based agent, while figures \ref{fig:time_tnr} and \ref{fig:time_aql} display the outcomes for the time-based agent. Early in training, fluctuations in rewards and queue lengths are observed due to high exploration rates. As training progresses and the agent optimizes its policy, rewards increase and queue lengths decrease, indicating improved performance. By the end of training, both metrics stabilize, confirming the agent's convergence and the effectiveness of the learning process.

%
%
%
\newcolumntype{M}[1]{>{\centering\arraybackslash}m{#1}}
\begin{table*}
    \centering
    \caption{Turn-based agent performance metrics}
    \begin{tabular}{M{7em} | M{5em} M{2em} M{5em} M{2em} M{5em} M{2em} M{5em} M{2em}}
    \toprule
    \textbf{Performance} & \multicolumn{8}{c}{\textbf{Traffic Scenarios}} \\ 
    \cmidrule{2-9}
    \textbf{Measure} & \textbf{Low} & \textbf{\%} & \textbf{High} & \textbf{\%} & \textbf{EW} & \textbf{\%} & \textbf{NS} & \textbf{\%} \\
    \midrule
    $Tnr$ & -3046.4\cellcolor[HTML]{F2F2F2} & 38\cellcolor[HTML]{EAF1DD} & -100468.2\cellcolor[HTML]{F2F2F2} & -23\cellcolor[HTML]{F2DBDB} & -7244.2\cellcolor[HTML]{F2F2F2} & 73\cellcolor[HTML]{EAF1DD} & -6979.6\cellcolor[HTML]{F2F2F2} & 69\cellcolor[HTML]{EAF1DD} \\[2pt]
    $Tawt$ & 13097.2\cellcolor[HTML]{F2F2F2} & 36\cellcolor[HTML]{EAF1DD} & 6016158.8\cellcolor[HTML]{F2F2F2} & -67\cellcolor[HTML]{F2DBDB} & 42486\cellcolor[HTML]{F2F2F2} & 62\cellcolor[HTML]{EAF1DD} & 41651.4\cellcolor[HTML]{F2F2F2} & 53\cellcolor[HTML]{EAF1DD} \\ [2pt]
    $ewpv$ & 14.0773\cellcolor[HTML]{F2F2F2} & 38\cellcolor[HTML]{EAF1DD} & 178.8351\cellcolor[HTML]{F2F2F2} & -17\cellcolor[HTML]{F2DBDB} & 15.9713\cellcolor[HTML]{F2F2F2} & 56\cellcolor[HTML]{EAF1DD} & 15.8665\cellcolor[HTML]{F2F2F2} & 51\cellcolor[HTML]{EAF1DD} \\ [2pt]
    $aql\ (E)$ & 0.5081\cellcolor[HTML]{F2F2F2} & 46\cellcolor[HTML]{EAF1DD} & 27.4442\cellcolor[HTML]{F2F2F2} & -9\cellcolor[HTML]{F2DBDB} & 1.8095\cellcolor[HTML]{F2F2F2} & 67\cellcolor[HTML]{EAF1DD} & 1.1115\cellcolor[HTML]{F2F2F2} & -40\cellcolor[HTML]{F2DBDB} \\ [2pt]
    $aql\ (W)$ & 0.4971\cellcolor[HTML]{F2F2F2} & 39\cellcolor[HTML]{EAF1DD} & 26.6061\cellcolor[HTML]{F2F2F2} & -13\cellcolor[HTML]{F2DBDB} & 1.9008\cellcolor[HTML]{F2F2F2} & 64\cellcolor[HTML]{EAF1DD} & 1.0488\cellcolor[HTML]{F2F2F2} & -8\cellcolor[HTML]{F2DBDB} \\ [2pt]
    $aql\ (N)$ & 0.5826\cellcolor[HTML]{F2F2F2} & 36\cellcolor[HTML]{EAF1DD} & 27.3906\cellcolor[HTML]{F2F2F2} & -7\cellcolor[HTML]{F2DBDB} & 1.0860\cellcolor[HTML]{F2F2F2} & -31\cellcolor[HTML]{F2DBDB} & 1.9620\cellcolor[HTML]{F2F2F2} & 62\cellcolor[HTML]{EAF1DD} \\ [2pt]
    $aql\ (S)$ & 0.5433\cellcolor[HTML]{F2F2F2} & 33\cellcolor[HTML]{EAF1DD} & 27.0814\cellcolor[HTML]{F2F2F2} & -21\cellcolor[HTML]{F2DBDB} & 1.0732\cellcolor[HTML]{F2F2F2} & -9\cellcolor[HTML]{F2DBDB} & 1.8245\cellcolor[HTML]{F2F2F2} & 60\cellcolor[HTML]{EAF1DD} \\ [2pt]
    \bottomrule
    \end{tabular}
    \label{tab:turn_result}
\end{table*}
\begin{table*}
    \centering
    \caption{Time-based agent performance metrics}
    \begin{tabular}{M{7em} | M{5em} M{2em} M{5em} M{2em} M{5em} M{2em} M{5em} M{2em}}
    \toprule
    \textbf{Performance} & \multicolumn{8}{c}{\textbf{Traffic Scenarios}} \\
    \cmidrule{2-9}
    \textbf{Measure} & \textbf{Low} & \textbf{\%} & High & \textbf{\%} & \textbf{EW} & \textbf{\%} & \textbf{NS} & \textbf{\%}\\
    \midrule
    $Tnr$ & -4387.4\cellcolor[HTML]{F2F2F2} & 11\cellcolor[HTML]{EAF1DD} & -45507\cellcolor[HTML]{F2F2F2} & 44\cellcolor[HTML]{EAF1DD} & -15091.2\cellcolor[HTML]{F2F2F2} & 43\cellcolor[HTML]{EAF1DD} & -14955.8\cellcolor[HTML]{F2F2F2} & 34\cellcolor[HTML]{EAF1DD} \\[2pt]
    $Tawt$ & 16366.6\cellcolor[HTML]{F2F2F2} & 20\cellcolor[HTML]{EAF1DD} & 983491.4\cellcolor[HTML]{F2F2F2} & 73\cellcolor[HTML]{EAF1DD} & 55666.8\cellcolor[HTML]{F2F2F2} & 50\cellcolor[HTML]{EAF1DD} & 54893.6\cellcolor[HTML]{F2F2F2} & 38\cellcolor[HTML]{EAF1DD} \\[2pt]
    $ewpv$ & 18.077\cellcolor[HTML]{F2F2F2} & 20\cellcolor[HTML]{EAF1DD} & 83.6828\cellcolor[HTML]{F2F2F2} & 45\cellcolor[HTML]{EAF1DD} & 23.0779\cellcolor[HTML]{F2F2F2} & 37\cellcolor[HTML]{EAF1DD} & 22.8631\cellcolor[HTML]{F2F2F2} & 30\cellcolor[HTML]{EAF1DD} \\[2pt]
    $aql\ (E)$ & 0.72\cellcolor[HTML]{F2F2F2} & 23\cellcolor[HTML]{EAF1DD} & 7.9955\cellcolor[HTML]{F2F2F2} & 68\cellcolor[HTML]{EAF1DD} & 3.3553\cellcolor[HTML]{F2F2F2} & 38\cellcolor[HTML]{EAF1DD} & 0.6877\cellcolor[HTML]{F2F2F2} & 14\cellcolor[HTML]{EAF1DD} \\[2pt]
    $aql\ (W)$ & 0.695\cellcolor[HTML]{F2F2F2} & 15\cellcolor[HTML]{EAF1DD} & 10.3135\cellcolor[HTML]{F2F2F2} & 56\cellcolor[HTML]{EAF1DD} & 3.4459\cellcolor[HTML]{F2F2F2} & 35\cellcolor[HTML]{EAF1DD} & 0.7855\cellcolor[HTML]{F2F2F2} & 19\cellcolor[HTML]{EAF1DD} \\[2pt]
    $aql\ (N)$ & 0.697\cellcolor[HTML]{F2F2F2} & 24\cellcolor[HTML]{EAF1DD} & 9.3785\cellcolor[HTML]{F2F2F2} & 63\cellcolor[HTML]{EAF1DD} & 0.7070\cellcolor[HTML]{F2F2F2} & 15\cellcolor[HTML]{EAF1DD} & 3.2986\cellcolor[HTML]{F2F2F2} & 36\cellcolor[HTML]{EAF1DD} \\[2pt]
    $aql\ (S)$ & 0.658\cellcolor[HTML]{F2F2F2} & 19\cellcolor[HTML]{EAF1DD} & 11.2018\cellcolor[HTML]{F2F2F2} & 50\cellcolor[HTML]{EAF1DD} & 0.7615\cellcolor[HTML]{F2F2F2} & 23\cellcolor[HTML]{EAF1DD} & 3.4375\cellcolor[HTML]{F2F2F2} & 25\cellcolor[HTML]{EAF1DD} \\[2pt]
    \bottomrule
    \end{tabular}
    \label{tab:time_result}
\end{table*}
%
%
%
\begin{figure*}
    \centering
    \subfloat[Turn-based agent turn count\label{fig:turn_count}]{%
       \includegraphics[width=0.3\linewidth]{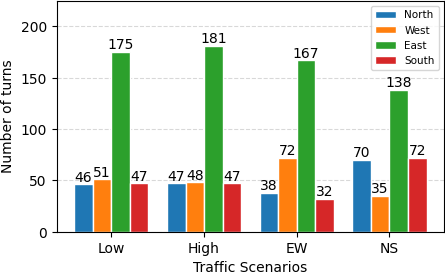}}
    \hfill
    \subfloat[Queue length plot for turn-based agent\label{fig:turn_ql}]{%
       \includegraphics[width=0.29\linewidth]{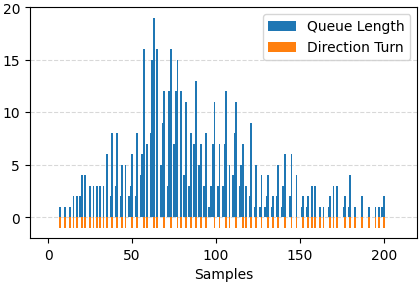}}
    \hfill
    \subfloat[Queue length plot for time-based agent\label{fig:time_ql}]{%
        \includegraphics[width=0.3\linewidth]{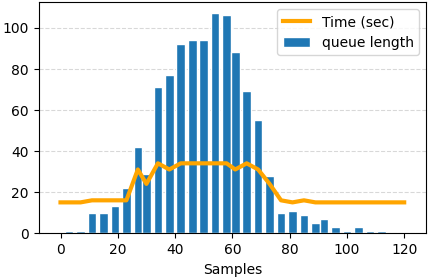}}
    \caption{Actions performed by agent during one episode.}
    \label{fig:action}
\end{figure*}
%
%
%
\begin{figure*}
    \centering
    \subfloat[Performance comparison on each traffic scenario\label{fig:pef_each_sc}]{%
       \includegraphics[width=0.3\linewidth]{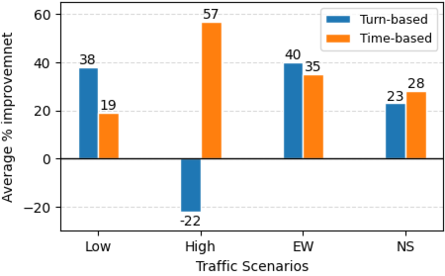}}
    \hspace{20mm}
    \subfloat[Average performance over all traffic scenarios\label{fig:avg_pef}]{%
        \includegraphics[width=0.3\linewidth]{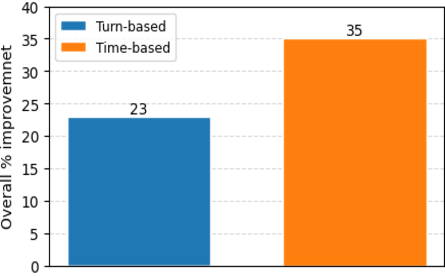}}
    \caption{Performance comparison plot of turn based agent and time based agent.}
    \label{fig:comp_result}
\end{figure*}
\subsection{Results}
To assess the performance of the proposed agents, we compared them against a conventional traffic light control system. The evaluation involved running five episodes of SUMO simulation with randomly generated traffic under four fixed scenarios: low, high, East-West (\textit{EW}), and North-South (\textit{NS}). Seven evaluation metrics were computed for each episode, and their mean values were calculated to determine the average performance for each metric. This procedure was applied to both reinforcement learning agents and the conventional control system to gauge performance improvements.

\subsubsection{Turn-based Agent}
Table \ref{tab:turn_result} presents the evaluation results for the turn-based agent across the four traffic scenarios. The columns labeled with traffic scenarios show the metric outcomes, while the percentage improvement columns, highlighted in green and red, indicate performance changes relative to the conventional system. Green signifies improvement, while red indicates a decline in performance. 

The turn-based agent demonstrated strong performance in low traffic scenarios but struggled in high traffic conditions. In the \textit{EW} and \textit{NS} scenarios, the agent displayed moderate performance. The agent's behavior is characterized by a tendency to prioritize sides with higher queue lengths, which is beneficial in certain conditions but problematic in high traffic scenarios. Here, the agent frequently switches phases due to uniformly high queues in all directions, resulting in excessive transitions and increased vehicle wait times. This behavior reduces overall efficiency, as reflected in the poor performance metrics.

Fig. \ref{fig:turn_count} illustrates the turn count distribution across all scenarios, showing a preference for the east direction. This pattern suggests the agent defaults to east when traffic is minimal, as seen at the beginning and end of simulations. In balanced scenarios, the agent makes fairly even decisions, except for a bias toward east. In directional scenarios like \textit{NS} and \textit{EW}, the agent appropriately favors the direction with more traffic.

Fig. \ref{fig:turn_ql} shows the queue length over time for a low traffic scenario. Positive values (blue) indicate queue length, while negative values (orange) represent the sampling timesteps at which the agent took action. The data suggests that the agent acts primarily when vehicles are present, minimizing unnecessary phase changes.

\subsubsection{Time-based Agent}
The time-based agent was designed to address the shortcomings of the turn-based agent, particularly its greediness in favoring high-queue directions. By cycling through all phases regardless of queue lengths, the time-based agent ensures that even sides with shorter queues are periodically served.

Table \ref{tab:time_result} details the performance of the time-based agent under different traffic conditions. The results indicate consistent improvement across all scenarios compared to the conventional system, with all percentage improvement columns showing positive values. The agent's robust performance is attributed to its adaptive phase duration, which adjusts based on real-time traffic conditions. In low traffic scenarios, it uses minimal phase times, whereas in high traffic scenarios, it extends phase durations up to the maximum. This adaptability allows the agent to handle varying traffic loads effectively.

Fig. \ref{fig:time_ql} shows the queue length and phase duration over one episode. The blue bars represent queue length, while the orange line shows phase duration. The Fig. demonstrates that as queue length increases, the agent responds with longer phase durations, reaching up to $34$ seconds when necessary. When traffic is low, phase durations reduce back to the minimum of $15$ seconds, illustrating the agent's capacity to learn and apply an effective policy.

\subsubsection{Comparison}
To determine the superior agent, we compared the performance of the turn-based and time-based agents across all scenarios. By averaging the percentage improvements across all seven evaluation metrics, we derived a single scalar value representing each agent's overall performance enhancement over the conventional system.

Fig. \ref{fig:pef_each_sc} displays the results for both agents across the four traffic scenarios. The vertical axis represents the average percentage improvement, while the horizontal axis denotes the traffic scenarios. The turn-based agent outperforms the time-based agent in low traffic scenarios due to its focused phase allocation, which reduces unnecessary signaling. However, in high traffic conditions, the time-based agent excels by dynamically adjusting phase durations, mitigating the impact of high traffic volumes more effectively than the turn-based agent. In directional scenarios like \textit{EW} and \textit{NS}, the agents' performances are comparable, each showing strengths in different contexts.

To identify the overall best-performing agent, we calculated the mean of the average percentage improvements across all scenarios (Fig. \ref{fig:avg_pef}). The time-based agent emerges as the superior option, with a $35\%$ average improvement across all traffic scenarios, compared to the turn-based agent. 

The findings from this study highlight the potential of reinforcement learning in adaptive traffic control. The choice between a turn-based and time-based agent should depend on the typical traffic conditions encountered at a given intersection. For intersections frequently experiencing high traffic, a time-based agent is preferable due to its substantial performance gains (up to $57\%$). For intersections with lower or varied traffic, the turn-based agent can perform adequately. However, for intersections facing diverse traffic conditions, the time-based agent is more reliable due to its consistent performance.
\section{Conclusion}
In this study, we introduced two reinforcement learning agents—turn-based and time-based—to address traffic congestion at intersections. Both agents employ a novel feature encoding approach that transforms scalar queue length inputs into a feature matrix, enabling the discovery of optimal traffic signal control policies. Unlike traditional approaches that rely solely on a reward function for performance assessment, we utilized a comprehensive set of seven evaluation metrics to quantify performance improvements.

Simulation results demonstrate that both agents significantly outperform conventional traffic light control systems across various traffic scenarios. A comparative analysis revealed that the turn-based agent is particularly effective under low traffic conditions, while the time-based agent excels in high traffic scenarios. Furthermore, an aggregated evaluation across all scenarios indicates that the time-based agent consistently delivers superior performance, achieving an average improvement over the turn-based agent.

These findings underscore the potential of reinforcement learning for adaptive traffic control, suggesting that time-based strategies may offer more robust solutions for intersections experiencing a wide range of traffic conditions.

\bibliographystyle{IEEEtran}
\bibliography{references}

\end{document}